\documentclass{article}

\usepackage{microtype}
\usepackage{graphicx}
\usepackage{subfigure}
\usepackage{booktabs} 
\usepackage{multirow}

\usepackage{hyperref}

\usepackage[accepted]{icml2021}

\icmltitlerunning{IowaRain: A Statewide Rain Event Dataset Based on Weather Radars and Quantitative Precipitation Estimation}

\begin{document}

\twocolumn[
\icmltitle{IowaRain: A Statewide Rain Event Dataset Based on Weather Radars and Quantitative Precipitation Estimation}



\icmlsetsymbol{equal}{*}

\begin{icmlauthorlist}
\icmlauthor{Muhammed Sit}{iihr}
\icmlauthor{Bong-Chul Seo}{iihr}
\icmlauthor{Ibrahim Demir}{iihr}

\end{icmlauthorlist}

\icmlaffiliation{iihr}{IIHR—Hydroscience \& Engineering, The University of Iowa, Iowa City, Iowa, USA}

\icmlcorrespondingauthor{Muhammed Sit}{muhammed-sit@uiowa.edu}

\icmlkeywords{Environment, Benchmark Dataset, Rainfall, Precipitation}

\vskip 0.3in
]



\printAffiliationsAndNotice{}  

\begin{abstract}
Effective environmental planning and management to address climate change could be achieved through extensive environmental modeling with machine learning and conventional physical models. In order to develop and improve these models, practitioners and researchers need comprehensive benchmark datasets that are prepared and processed with environmental expertise that they can rely on. This study presents an extensive dataset of rainfall events for the state of Iowa (2016-2019) acquired from the National Weather Service Next Generation Weather Radar (NEXRAD) system and processed by a quantitative precipitation estimation system. The dataset presented in this study could be used for better disaster monitoring, response and recovery by paving the way for both predictive and prescriptive modeling.

\end{abstract}

\section{Introduction}
\label{intro}

Many problems in the environmental domain can be tackled with data-driven approaches. However, as with any data-driven problem, the lack of available data limits the development and applicability of environmental modeling efforts. Furthermore, environmental modeling, specifically probability-based environmental modeling, in contrast with its dependency on data, suffers from data scarcity \cite{sit2020comprehensive}. The data scarcity problem in the field is being iterated in many studies \cite{ebert2017vision, rolnick2019tackling} while the cardinality of publicly available pre-processed datasets remains to be low.

\subsection{Related Work}

Studies rectify data scarcity one step at a time. CAMELS being one of the oldest and widely used datasets in rainfall-runoff prediction tasks, provides a go-to mark for practitioners working on flood forecasting for contagious United States \cite{newman2015development}. CAMELS provides streamflow measurements from the nation-wide United States Geological Survey (USGS) river sensor network as well as essential data points that are typically used in rainfall-runoff modeling. Although CAMELS is not prepared with statistical predictive modeling in mind to support physical modeling, another study, FlowDB \cite{godfried2020flowdb} presents two extensive datasets with a machine learning focus. The first dataset, following in CAMEL’s footsteps, focuses on streamflow levels along with precipitation measurements for flood forecasting purposes. FlowDB also goes beyond streamflow with the second dataset on flash flood events. The second dataset includes many aspects of historic flood events, including river flow, precipitation measurements, property damage estimations, injury data, and event locations. While both CAMELS and FlowDB involve precipitation measurements, their main focus falls into the flood forecasting domain rather than pure rainfall and precipitation context. 

One benchmark dataset focusing on rainfall is WeatherBench \cite{rasp2020weatherbench} which presents a dataset with a focus on medium-range weather forecasting pre-processed for deep learning applications. WeatherBench uses measurements from ERA5 numerical reanalysis dataset \cite{hersbach2020era5}. ERA5 combines a forecast model with the available observations and provides estimations of the atmospheric state. An alternative dataset using ERA5 is RainBench \cite{de2020rainbench}. In addition to numerical estimations from ERA5, RainBench also incorporates simulated satellite data as imagery and global precipitation estimates from IMERG\cite{huffman2015nasa} to support easier data-driven weather forecasting.

Even though the number of datasets published directly or remotely related to precipitation has been gaining momentum for the last few years, the need for alternative environmental monitoring datasets in order to ensure better climate change modeling and planning is critical \cite{rolnick2019tackling, sit2020comprehensive}. This study presents a rain event dataset consisting of precipitation data between 2016 and 2019 from the state of Iowa based on the NEXRAD radar network from seven radars positioned in and around the state. The main goal of this dataset is to support climate change modeling for effective disaster planning, response, and recovery. While the dataset presented here does not necessarily define a problem, it invites machine learning researchers to better study precipitation and rainfall by providing the dataset and various sample problems.

This paper is structured as follows; the introduction section briefly presents the problem that this paper tackles and covers previous works. The following section (Dataset) discusses the specifics of the dataset by describing how the raw data was acquired and processed. Then it shares descriptive statistics representing the dataset to form an understanding while listing potential environmental domain problems for that the dataset can be used. Lastly, in the conclusion section, we share the final remarks.

\begin{figure}[!htb]
\vskip 0.1in
\begin{center}
\centerline{\frame{\includegraphics[width=\columnwidth]{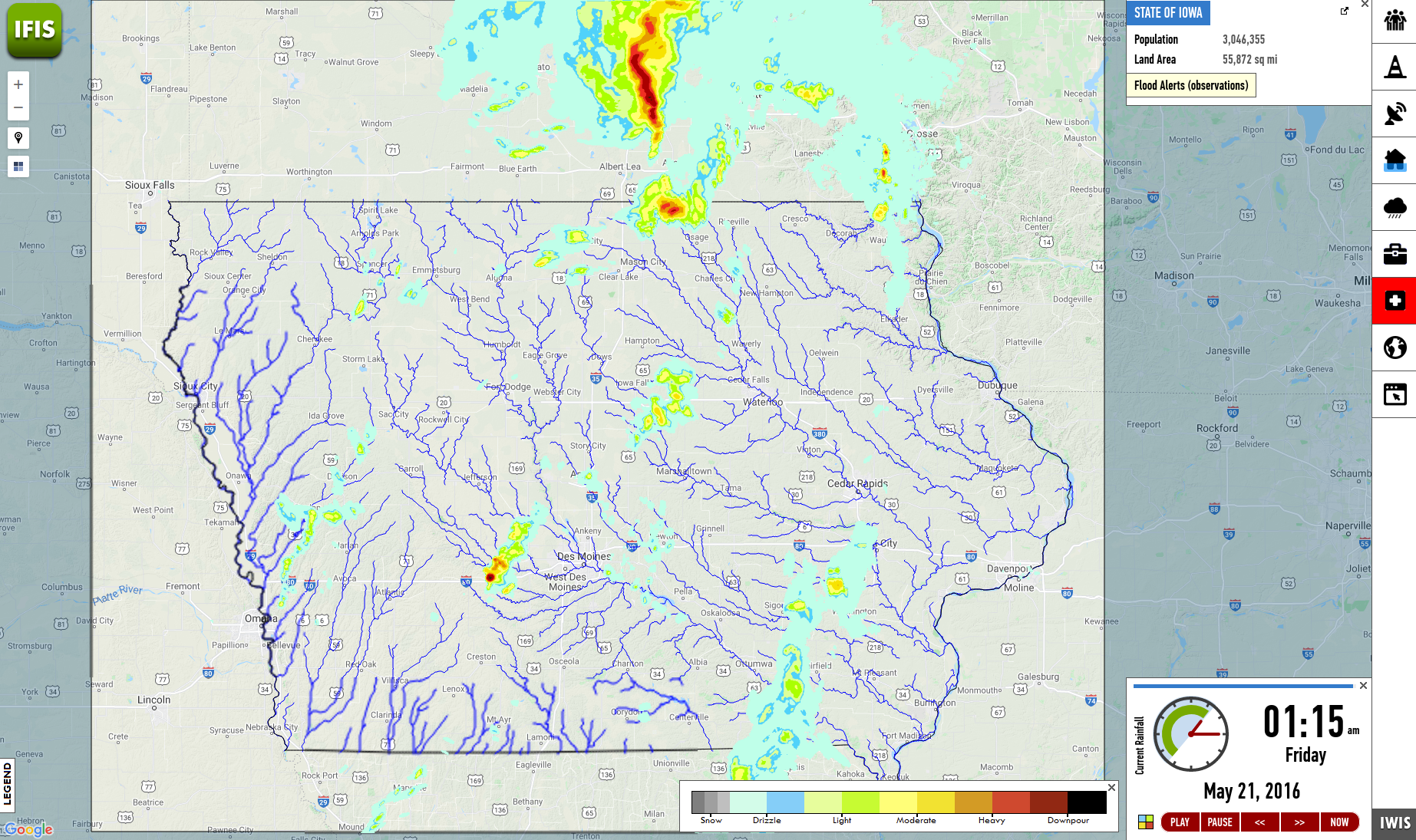}}}
\caption{A 2D rain rate map visualized for Iowa domain.}
\label{ifis}
\end{center}
\vskip -0.3in
\end{figure}

\begin{figure}[!htb]
\vskip 0.1in
\begin{center}
\centerline{\frame{\includegraphics[width=\columnwidth]{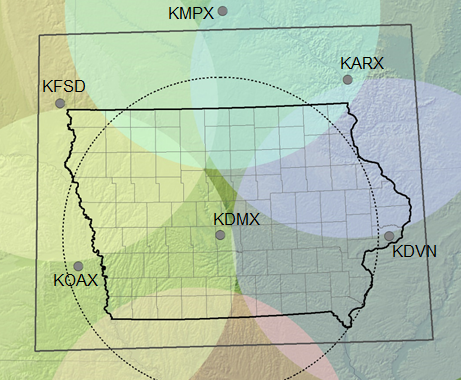}}}
\caption{WSR-88D network radar coverages over Iowa domain.} 
\label{wsr}
\end{center}
\vskip -0.2in
\end{figure}

\section{Dataset}

The IowaRain dataset comprises rainfall events from 2016 through the end of 2019 based on Iowa Flood Center's (IFC) Quantitative Precipitation Estimation (QPE) system. The IFC QPE system, for the most part, relies on the National Weather Service's Weather Surveillance Radar-1988 Doppler (WSR-88D) network, which is operational at over 160 locations in the United States. WSR-88Ds complete a scan every 4-5 minutes for approximately 80 nautical miles (nm) range and produce data representing precipitation within their respective domains. The generated QPE products are fed to the Rainfall Event Detection system to determine and separate rainfall events.

\begin{figure*}[ht]
\vskip 0.2in
\begin{center}
\centerline{\includegraphics[height=6cm,keepaspectratio]{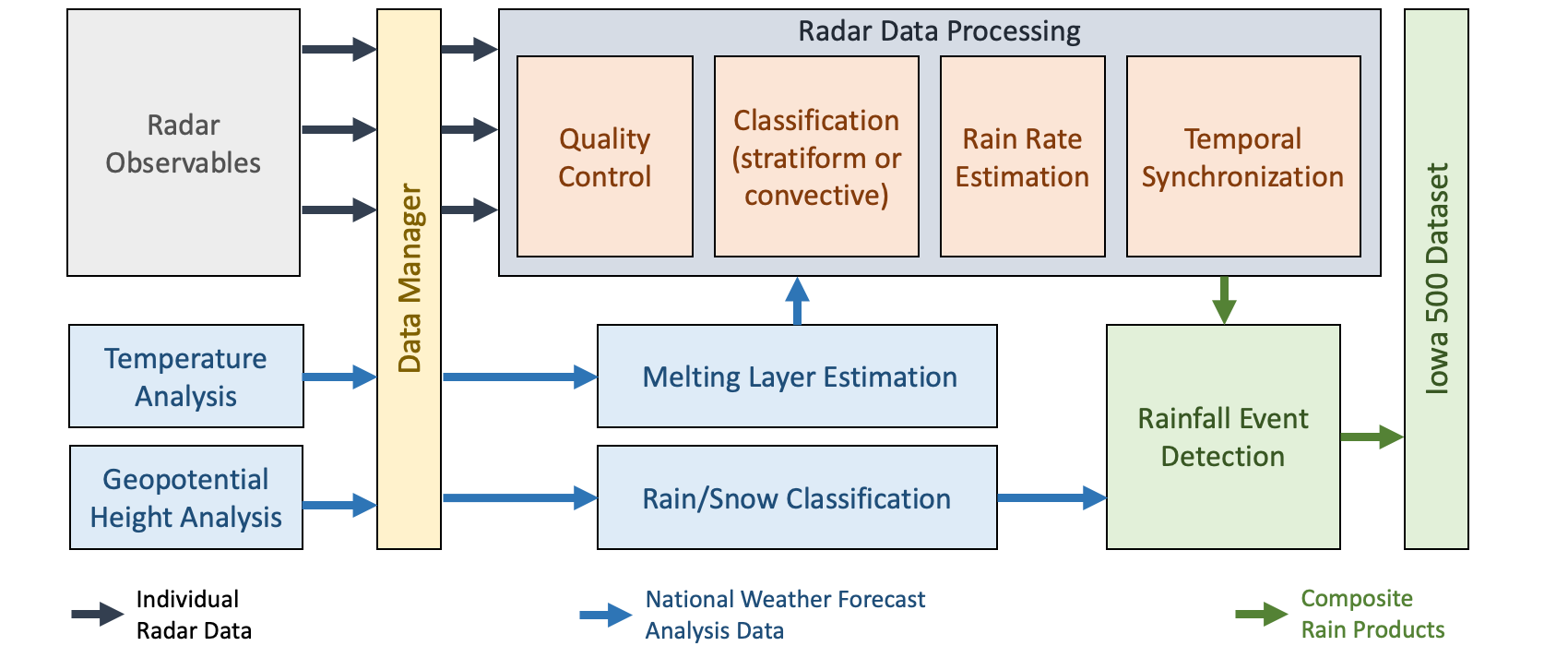}}
\caption{Overall summary of dataset acquisition and processing workflow.}
\label{arch}
\end{center}
\vskip -0.2in
\end{figure*}

The IFC QPE system \cite{seo2019pilot} uses raw data acquired in real-time from seven WSR-88D radars (KARX in La Crosse, Wisconsin; KDMX in Des Moines, Iowa; KDVN in Davenport, Iowa; KEAX in Kansas City, Missouri; KFSD in Sioux Falls, South Dakota; KMPX in Minneapolis, Minnesota; and KOAX in Omaha, Nebraska). In addition, the system incorporates the specific attenuation algorithm \cite{seo2020evaluation} to ensure the efficient employment of WSR 88D’s dual-polarization functions to improve QPE accuracy. To do so, it also determines melting layer locations by using temperature soundings from the numerical weather prediction models.

\subsection{QPE System \& Rainfall Event Detection}

Using the WSR-88D network, the QPE system generates a composite rain rate map (Figure \ref{ifis}) covering the IFC domain that bounds the state of Iowa with buffer zones encircling the state (Figure \ref{wsr}), with 5 minutes of temporal and 500 meters of spatial resolutions, using a variety of processing algorithms. The unit of the product is millimeters/hour (mm/hr).

These QPE products are matrices, named snapshots, with the shape of 1057 x 1741 for each timestamp in [2016-2019] that have their minute component divisible by 5. Even though the QPE system produces maps for snow as well, the IowaRain dataset makes a distinction and only comprises rainfall events from April to October each year to avoid any potential derangement.

Data acquisition and processing for IowaRain are summarized in (Figure \ref{arch}). The data manager module of the QPE system handles the acquisition of radar observables (reflectivity, radial velocity, spectrum, width, differential reflectivity, copular correlation coefficient, and differential phase) as well as temperature and geo-potential height analysis for various pressure levels from numerical weather prediction models. After the data acquisition, the Data Manager ensures that the data is complete.

Once the data is ready, it goes into a series of processing sub-modules within the Data Processing subsystem for quality control, rain rate estimation, merging of individual radar data, and finally, rainfall product generation. In the quality control step, the system essentially eliminates non-meteorological data radars acquired, i.e., noise. After the data quality check, the QPE system generates temperature soundings for areas that radars cover by utilizing the numerical weather prediction analysis and calculates the altitude of the melting layer using 3D spherical coordinates. The system then runs a series of algorithms to classify individual radar scans into one of convective and stratiform using melting layer information.

The melting layer and classification of precipitation are then used by an estimator to generate 2D rain rate maps separate for each elevation. These individual elevation maps then get combined by a non-parametric kernel function to build the final 2D product for each radar scan. These 2D products built from individual radars need to be combined into one 2D product for the Iowa domain at the last step. Beyond the straightforward combination of various 2D maps into one, this step needs spatial and temporal synchronization as radars differ from each other with their spatial and temporal coverages. By combining 2D radar scans according to characteristics of the radar they originate from, the QPE system produces a composite 2D rain rate map every 5 minutes. The details regarding the QPE system and discussion about its accuracy can be reviewed further in \cite{seo2020statewide}.
The composite rain rate maps are plain data files that need serialization for easier usage in data-driven tasks. To facilitate that, the Rainfall Detection System is used to read all the 2D plain rain rate maps into NumPy \cite{harris2020array} arrays and then to be normalized by dividing them to 105, which records the presumably maximum observation in the Iowa domain. Since the matrices/snapshots the QPE system produces have odd shapes, we resized them to (1088, 1760) by adding empty elements to enable more straightforward utilization of architectures that use iterative downsampling. This design choice could easily be reverted by removing extra rows and columns.

In order to create the final dataset, the serialized snapshots go through rainfall detection criteria. The criteria for a set of snapshots are defined as follows;
\begin{itemize}
\itemsep0em
  \item All snapshots in the set must be consecutive
  \item The set must consist of at least ten snapshots
  \item Each snapshot must have at least 0.5 mm/hr precipitation over \%10 of the domain.
\end{itemize}

The final version of the dataset consists of gzip files, one archive for each rain event. When any of the gzip archives is opened in Python with pickle, each file reads a triple that is structured as $(start\_date, number\_of\_snapshots, list\_of\_snapshots)$. The IowaRain dataset described in this section and sample code to open individual rainfall events can be found in the Git repository at \href{https://github.com/uihilab/IowaRain}{https://github.com/uihilab/IowaRain}.

\begin{table}[!htb]
\caption{Number of events and average event duration by year.}
\label{number-by-year}
\vskip 0.15in
\begin{center}
\begin{small}
\begin{sc}
\begin{tabular}{lcccr}
\toprule
Year & Number of Events & Avg Event Length \\
\midrule
2016    & 64 & 7.09 Hours \\
2017    & 67 & 6.55 Hours\\
2018    & 76 & 8.04 Hours \\
2019    & 81 & 7.09 Hours \\
\bottomrule
\end{tabular}
\end{sc}
\end{small}
\end{center}
\vskip -0.1in
\end{table}

\subsection{Dataset Statistics}
The IowaRain dataset comprises 1242 rainfall events with various durations. Since each year has different weather characteristics, there is no normal distribution of rainfall events year by year. While there is an increasing trend, the number of events each year is not drastically different (Table \ref{number-by-year}). The length of the longest rain event in 2019 is lower than the previous year by a large margin, while the minimum event length stays the same per the aforementioned event criteria (Table \ref{max-min}). A summary of events can be seen in Table \ref{coverage-prec}, where we present average coverage of the domain and average rain rate in mm/hr by month and year.

\begin{table}
\caption{Shortest and longest rainfall events by year.}
\label{max-min}
\vskip 0.15in
\begin{center}
\begin{small}
\begin{sc}
\begin{tabular}{lcccr}
\toprule
Year & Min Event Length & Max Event Length \\
\midrule
2016    & 50 Minutes & 29 Hours \\
2017    & 50 Minutes & 29 Hours, 15 Minutes\\
2018    & 50 Minutes & 33 Hours, 55 Minutes \\
2019    & 50 Minutes & 17 Hours, 40 Minutes \\
\bottomrule
\end{tabular}
\end{sc}
\end{small}
\end{center}
\vskip -0.1in
\end{table}

\begin{table}[H]
\caption{Average rainfall coverage and average precipitation of events by year and month.}
\label{coverage-prec}
\vskip 0.15in
\begin{center}
\begin{small}
\begin{sc}
\begin{tabular}{lccccr}
\toprule
Year & Month & Avg Coverage & Avg Precipitation \\
\midrule
\multirow{7}{*}{2016}&Apr & 16.81\% & 0.6370\\
	&May & 15.18\% & 0.6741\\
	&Jun & 12.90\% & 0.9355\\
	&Jul & 17.12\% & 1.3949\\
	&Aug & 13.56\% & 1.0103\\
    &Sep & 16.05\% & 1.0858\\
    &Oct & 17.25\% & 0.8573\\
\multirow{7}{*}{2017}&Apr & 14.01\% & 0.5657\\
    &May & 16.50\% & 0.7977\\
    &Jun & 13.00\% & 1.0379\\
    &Jul & 14.11\% & 0.9905\\
    &Aug & 14.16\% & 0.9263\\
    &Sep & 14.56\% & 0.6125\\
    &Oct & 14.32\% & 0.6791\\
\multirow{7}{*}{2018}&Apr & 14.60\% & 0.3699\\
    &May & 17.01\% & 0.9445\\
    &Jun & 14.94\% & 0.9249\\
    &Jul & 12.93\% & 0.8124\\
    &Aug & 14.04\% & 0.7316\\
    &Sep & 16.63\% & 0.9452\\
    &Oct & 13.80\% & 0.4725\\
\multirow{7}{*}{2019}&Apr & 15.26\% & 0.5559\\
    &May & 16.88\% & 0.7453\\
    &Jun & 14.47\% & 0.6307\\
    &Jul & 11.94\% & 0.8102\\
    &Aug & 15.34\% & 1.0646\\
    &Sep & 16.59\% & 1.0869\\
    &Oct & 17.93\% & 0.6226\\

\bottomrule
\end{tabular}
\end{sc}
\end{small}
\end{center}
\vskip -0.1in
\end{table}

Even though in data-driven environmental modeling, train/validation/test split is typically done by date like many other time-series focused problems, we do not necessarily put a constraint on that in this study. Since each rainfall event carries different characteristics, we leave the dataset splitting to other researchers' preferences.

\subsection{Sample Tasks}

It is needless to say that weather and rainfall forecasting typically are done by using more than just previous precipitation measurements. Nevertheless, IowaRain presents a unique challenge in rainfall forecasting, that is, forecasting rainfall events both short-term and long-term with limited data access. This problem is very much like the video frame prediction task, and since frame prediction is a well-studied problem, extensions of the state-of-the-art architectures for the video frame prediction problem could be explored with IowaRain.

Another potential task that might be traversed is about the dataset’s nature. For example, as IowaRain is based on the WSR-88D network, it is prone to artifacts caused by the noise that is confusing radars. Those artifacts could be identified, and snapshots with artifacts could be labeled as an extension, and a data cleaning focused unsupervised approach could remove artifacts from snapshots. 

One other outlook for IowaRain focuses on streamflow. Rainfall-runoff forecasting is one of the most studied problems in hydroscience with both physical-based and machine learning models. IowaRain enables practitioners and researchers to study streamflow by combining IowaRain with publicly available USGS sensor networks. By incorporating these datasets together, one could work on streamflow forecasting for both gauged and ungauged locations.

\section{Conclusions}
This paper presents a rainfall event dataset, namely \href{https://github.com/uihilab/IowaRain}{IowaRain}, generated from weather radars and various other environmental analysis models. First, we briefly explained how the weather products from several radars covering areas in Iowa are acquired, processed, synchronized, and combined into 2D rain rate maps with temporal and spatial resolutions of 5 minutes and 500 meters, respectively. Subsequently, we described how individual rain events were determined and sampled how IowaRain could be helpful by outlining potential use cases as a starting point. Even though in its current form, the dataset covers only the state of Iowa and 2016-2019 as the timeframe, the same methodology described in this paper could be employed to extend the dataset for more comprehensive temporal and spatial coverage for the entire US.




\bibliography{main}
\bibliographystyle{icml2021}

\end{document}